\PassOptionsToPackage{dvipsnames}{xcolor}
\documentclass[11pt]{article}
\usepackage{textgreek}
\usepackage{amsmath}
\usepackage[final]{acl}
\usepackage{booktabs}
\usepackage{tikz}
\usepackage{booktabs,tabularx}
\usepackage{times}
\usepackage{latexsym}
\usepackage{booktabs}
\usepackage{multirow} 
\usepackage{hyperref}
\usepackage{url}
\usepackage{makecell}
\usepackage[T1]{fontenc}

\usepackage[utf8]{inputenc}
\usepackage{microtype}

\usepackage{inconsolata}

\usepackage{graphicx}

\usepackage{graphicx,booktabs}
\usepackage{tikz}

\newcommand{\maxval}{2.10}          
\newcommand{\barcell}[2]{
  \begin{tikzpicture}[x=\barwidth,y=0.22cm,baseline=0.2ex]
    \fill[black!12] (0,0) rectangle (1,1);   
    \fill[#2] (0,0) rectangle (#1/\maxval,1);
  \end{tikzpicture}\,\scriptsize #1\%%
}

\newcommand{\maxvals}{0.07}                         
\newcommand{\barcells}[2]{
  \begin{tikzpicture}[x=\barwidth,y=0.22cm,baseline=0.2ex]
    \fill[black!12] (0,0) rectangle (1,1);          
    \fill[#2] (0,0) rectangle (#1/\maxvals,1);       
  \end{tikzpicture}\,\scriptsize #1\%%
}

\newlength{\barwidth}
\newcommand{\maxtox}{3.2}    
\newcommand{\maxhate}{0.56}  

\newcommand{\barcelltox}[2]{
  \begin{tikzpicture}[x=\barwidth,y=0.22cm,baseline=0.2ex]
    \fill[black!12] (0,0) rectangle (1,1);
    \fill[#2] (0,0) rectangle (#1/\maxtox,1);
  \end{tikzpicture}\,\scriptsize #1\%%
}
\newcommand{\barcellhate}[2]{
  \begin{tikzpicture}[x=\barwidth,y=0.22cm,baseline=0.2ex]
    \fill[black!12] (0,0) rectangle (1,1);
    \fill[#2] (0,0) rectangle (#1/\maxhate,1);
  \end{tikzpicture}\,\scriptsize #1\%%
}

\title{Bias Dynamics in BabyLMs: Towards a Compute-Efficient Sandbox for Democratising Pre-Training Debiasing}

\author{
  \textbf{Filip Trhlik},
  \textbf{Andrew Caines},
  \textbf{Paula Buttery},
\\
    Department of Computer Science \& Technology,
    University of Cambridge, U.K.\\
    ALTA Institute,
    University of Cambridge, U.K.
\\
  \small{
    \href{mailto:ft360@cam.ac.uk}{\texttt{ft360@cam.ac.uk}}
  }
}

\begin{document}
\maketitle
\begin{abstract}
Pre-trained language models (LMs) have, over the last few years, grown substantially in both societal adoption and training costs. This rapid growth in size has constrained progress in understanding and mitigating their biases. Since re-training LMs is prohibitively expensive, most debiasing work has focused on post-hoc or masking-based strategies, which often fail to address the underlying causes of bias. In this work, we seek to democratise pre-model debiasing research by using low-cost proxy models. Specifically, we investigate BabyLMs, compact BERT-like models trained on small and mutable corpora that can approximate bias acquisition and learning dynamics of larger models. We show that BabyLMs display closely aligned patterns of intrinsic bias formation and performance development compared to standard BERT models, despite their drastically reduced size. Furthermore, correlations between BabyLMs and BERT hold across multiple intra-model and post-model debiasing methods. Leveraging these similarities, we conduct pre-model debiasing experiments with BabyLMs, replicating prior findings and presenting new insights regarding the influence of gender imbalance and toxicity on bias formation. Our results demonstrate that BabyLMs can serve as an effective sandbox for large-scale LMs, reducing pre-training costs from over 500 GPU-hours to under 30 GPU-hours. This provides a way to democratise pre-model debiasing research and enables faster, more accessible exploration of methods for building fairer LMs.

\end{abstract}

\section{Introduction}
The recent dramatic increase in investment in large language models (LLMs) has driven sharp performance gains \citep{10.1093/epolic/eiae057} while altering the way research is conducted. Since 2019, parameter counts have grown by roughly three orders of magnitude \cite{Radford2019LanguageMA, llama4blog}, notably inflating experimental costs and straining an academic ecosystem built on small, incremental advances \citep{cottier2024rising, sathish2024llempower}.

This problem is particularly pressing in LM bias research, which examines how LLMs treat users differently on the basis of protected attributes -- e.g., gender \citep{zhao2024gender} and ethnicity \citep{field-etal-2021-survey}. Despite the ubiquity of LMs, bias-removal research remains relatively linear and, constrained by the high cost of training, focuses mostly on post-hoc debiasing of already trained LMs, altering models only after their internal structures and reasoning circuits have formed \citep{gallegos2024bias}. While these methods can help, they often merely mask biases detectable by our simplistic probes rather than fully removing them \citep{gonen2019lipstick, gupta-etal-2024-sociodemographic}.

By contrast, debiasing strategies before or during pre-training remain rare and are outdated, partly because even a single pre-training run of a relatively simple BERT-style model exceeds 500 GPU hours on current hardware \citep{devlin2019bert}. This work investigates whether significantly less costly models could replace standard LMs in debiasing research, focusing on models from the BabyLM Challenge, a research initiative that seeks to create well-performing models trained on small-scale datasets \citep{warstadt-etal-2023-findings, charpentier2025babylm}. These properties make BabyLMs a promising democratisation tool, combining relevant performance that matches or surpasses BERT, mutable corpora that can be readily experimented with, and much lower pre-training time. In this paper, we verify this proposition, showing that BabyLMs:
\begin{enumerate}
\item \textit{acquire and express biases in a way that is representative of standard LMs}
\item \textit{respond to established debiasing methods similarly to standard LMs}
\item \textit{enable the democratisation of pre-model debiasing research}
\end{enumerate}

\section{Background}
\subsection{BabyLMs}
BabyLMs belong to the field of low-resource language models that seeks to democratise NLP with well-performing, affordable LMs \citep{van2024survey, conll-2023-babylm}. 

They are inspired by the fact that human children encounter three to four orders of magnitude less linguistic data than conventional LMs, with Llama-3 using approximately 15 trillion tokens \citep{grattafiori2024llama} while a 13-year-old child may only have encountered about 100 million words \citep{Gilkerson2017}. Unlike approaches that reduce parameter counts \citep{hoffmann2022training}, the BabyLM task restricts the amount of pre-training data, simulating a more human-like learning environment.

It utilises the aforementioned 100 million words as the training corpus, which contains transcriptions of child-directed speech, child-oriented texts (e.g., books, subtitles), and other commonly available data (e.g., Wikipedia) \cite{conll-2023-babylm}.

\subsection{Methods for Evaluating LM Performance}
\label{sec:Performance}
When introducing alternative LM architectures, we need evaluation frameworks to compare performance. \textbf{BLiMP} probes core grammatical knowledge via minimal pairs across multiple categories (syntax, semantics, etc.) \citep{warstadt-etal-2020-BLiMP-benchmark}. Additionally, \textbf{BabyLM BLiMP supplement} adds five extra dialogue/question probes \citep{warstadt-etal-2023-findings}. \textbf{EWoK} evaluates cognition-inspired world-model knowledge by asking models to match two contexts to two targets, discouraging reliance on surface likelihoods \citep{ivanova2024elements}. Lastly, more compute-intensive \textbf{GLUE/SuperGLUE} provide an overview of an LM's downstream capabilities~\citep{wang2018glue, wang2019superglue}.

\subsection{Notable BabyLM Architectures}
\label{sec:BabyLMs}
Throughout the years, numerous LM variants have been submitted to the competition, with the most relevant ones listed here. The \textbf{LTG-BERT} architecture \citep{samuel2023trained} keeps the BERT backbone but adds NormFormer \citep{shleifer2021normformer}, gated GELU \citep{shazeer2020glu}, disentangled relative positions \citep{he2020deberta}, and span masking \citep{joshi-etal-2020-spanbert}, outperforming BERT with a much smaller corpus for an identical number of training steps. \textbf{GPT-BERT} \citep{charpentier2024gpt}, the current SOTA BabyLM, retains LTG-BERT’s architecture but adds a hybrid objective that combines span masking with causal next-token prediction \citep{behnamghader2024llm2vec}, which further separates it from the classic BERT. Other approaches have also experimented with preprocessing \citep{cheng-etal-2023-mcgill} or curriculum learning \citep{martinez-etal-2023-climb}. However, they underperformed the LTG-based~architectures.

\subsection{Frameworks for Analysing LM Bias}
\label{sec:Bias}
\citet{blodgett-etal-2020-language} define bias as systematic patterns in representations or outputs that reinforce social inequalities, resulting in allocational or representational harms. Frameworks for evaluating LM bias range from intrinsic probes to more costly, task-specific extrinsic evaluations, with only a limited correlation observed between the two \citep{goldfarb2020intrinsic, cao-etal-2022-intrinsic}.

\textbf{StereoSet} measures intrinsic bias by making the model rank stereotypical, anti-stereotypical, and unrelated continuations~\citep{nadeem2020stereoset}. The bias score is derived from the ratio of examples in which the model prefers stereotypical over anti-stereotypical options, with an unbiased model yielding 50. Another notable dataset, \textbf{CrowS-Pairs} \citep{nangia2020crows}, uses entire stereotype/anti-stereotype sentence pairs, with the score again reflecting the proportion of stereotypical sentences chosen. Other frameworks include SEAT \citep{may-etal-2019-measuring}, which provides a lightweight intrinsic option, and WinoBias \citep{zhao2018gender}, which targets extrinsic coreference bias.

\subsection{Techniques for LM Bias Mitigation}
To examine techniques for reducing biases in LMs, we follow \citet{guo2024bias} and organise them by application stage. \textit{Post-model approaches}, which leave the model intact and adjust only representations, are cheap and interpretable but are surface-level: \textbf{Iterative Nullspace Projection} (INLP) removes linearly decodable protected-attribute signals \citep{ravfogel-etal-2020-null} and \textbf{Sent-Debias} projects away PCA-estimated bias subspaces \citep{liang-etal-2020-towards}. Nevertheless, non-linear probes frequently reveal residual bias, underscoring that post-hoc fixes are linear and incomplete \citep{sun2025aligned, gonen2019lipstick}.

\textit{Intra-model methods} involve fine-tuning that debiases the full model: increasing \textbf{dropout} can disrupt biased representations but risks negative performance impacts \citep{webster2020measuring}; \textbf{Counterfactual Data Substitution} (CDS) rewrites biased instances (e.g., swapping gendered entities), placing them into a new context \citep{bartl2020unmasking, webster-etal-2018-mind}; and \textbf{debiasing losses} that motivate the model to debias itself \citep{Park23Never}. While bringing some improvement, other studies have demonstrated the brittleness of these methods \citep{mendelson-belinkov-2021-debiasing}.

\textit{Pre-model methods} modify data or pre-training to reduce bias before training even starts. They yield more stable debiasing but are costly to implement and research due to the need to retrain the model \citep{li-etal-2024-data, xie2023empirical}. Key tactics include \textbf{Counterfactual Data Augmentation} (CDA) that swaps demographic markers to rebalance the entire training corpus \citep{lu2018gender, zmigrod-etal-2019-counterfactual, webster2020measuring}; \textbf{toxic-content filtering} that displays a beneficial debiasing effect when applied \citep{workshop2022bloom, ranaldi-etal-2024-trip}; and \textbf{perturbation augmentation}, which stochastically edits sentences along gender, ethnicity, and age axes to produce fairer models \citep{qian-etal-2022-perturbation}.

\section{Experiment Setup}
As established in the previous section, pre-model debiasing offers the most stable mitigation, addressing inner representations rather than surface cues. Yet these methods remain under-explored because they require costly re-training of a model from scratch and the manipulation of large, often improperly specified, corpora. Given the advantages of BabyLM architectures (competitive performance, compact datasets, and inexpensive pre-training), together with established bias- and performance-evaluation frameworks, we argue that BabyLMs can provide a practical sandbox for systematic pre-model debiasing and lower the research barrier.

To validate this proposition, we first need to understand how the debiasing behaviours of standard LMs and BabyLMs align. Therefore, we must identify candidate BabyLMs that suitably replicate the bias dynamics of standard models such as BERT.
\subsection{Metrics}
We utilise the performance and bias scores from frameworks described in Sections \ref{sec:Performance} and \ref{sec:Bias}. 
For bias, this means using CrowS-Pairs and StereoSet. Both share a mathematically similar approach and a score scale. Since BabyLMs can be either masked or continuation LMs and lack next-sentence prediction, we exclude StereoSet’s inter-sentence portion \citep{ranaldi-etal-2024-trip}.

A bias evaluation is not sufficient on its own. Since there is an established relationship between the model’s biases and its performance \citep{nadeem2020stereoset}, we estimate the performance through BLiMP, BabyLM BLiMP supplement, and EWoK. 

Thus, we obtain three performance metrics and two bias metrics, all of which capture only part of a model’s behaviour. They all share the same scale but probe the model with different sentences and contexts. Therefore, they reveal different parts of its bias and performance profile \citep{zakizadeh2025blind}. To obtain a more comprehensive picture, we average the individual performance scores into a \textit{composite performance} metric and the two bias scores into a \textit{composite bias} metric (metrics composition details in Appendix \ref{sec:align-models}).

\subsection{Candidate BabyLM}
As noted, our aim is to select a BabyLM that comes the closest behaviour-wise to standard LMs, while still retaining desirable characteristics, such as low-cost training. Firstly, this requires showing that BabyLMs in general display bias acquisition dynamics, such as verifying that they acquire more biases with increased performance, as observed within larger LMs \cite{nadeem2020stereoset}.

We do this by evaluating composite bias and performance metrics for every notable BabyLM and various variants of standard LMs, with all models used listed in Appendix~\ref{sec:eval-models}. This yields Table~\ref{tab:cor}, which shows the correlation between \textit{composite performance} and \textit{composite bias} for both model classes. The strong positive correlation, and its shared strength across BabyLMs and standard LMs, shows that the overall trend of bias increasing with performance is preserved in BabyLMs, with the exact spread of models shown in the Appendix \ref{sec:align-models}.  Therefore, BabyLMs can be a valid sandbox for studying debiasing techniques with resources that are feasible for many research groups.

\begin{table}[h!]
\centering
\begin{tabular}{l|c|c}
\toprule
\textbf{\shortstack[l]{Model\\Class}} & \textbf{$N$} &
\textbf{\shortstack{$r(\text{Composite Performance},$\\$\text{Composite Bias})$}} \\
\midrule
BabyLM   & 9  & 0.833 \\
Standard & 16 & 0.753 \\
\bottomrule
\end{tabular}
\caption{\label{tab:cor}Pearson correlation between composite performance and bias for the different model classes}
\vspace{-0.5em}
\end{table}

With BabyLM eligibility established, we select candidate models. The SOTA variant of the LTG-BERT architecture, \href{https://huggingface.co/ltg/ltg-bert-babylm}{\texttt{ltg-bert-babylm}}, is closest to the original BERT. However, it has been trained for 1,500 epochs, requiring roughly the same GPU-hours as the original BERT model, making it unsuitable for our purposes. Thus, we also select its low-resource variant, \href{https://huggingface.co/babylm/ltgbert-100m-2024}{\texttt{ltgbert-100m-2024}}, trained for just 40 GPU-hours. Although more distant from BERT, it still achieves reasonable performance and exhibits measurable bias, making it a promising low-cost BabyLM. In conclusion, the two candidate models for further study are \textbf{LTG-BERT} (ltg-bert-babylm) and \textbf{LTG-Baseline} (ltgbert-100m-2024). LTG-BERT tests whether any BabyLM can sufficiently replicate standard debiasing patterns, while LTG-Baseline tests whether these hold even with limited training.

\section{Model Viability}
Having established that BabyLMs acquire biases in a similar way to standard LMs, the next step is to test whether they also debias comparably. We analyse whether their pre-training corpora are comparable in terms of the biases acquirable from them. We also investigate how their composite performance and bias change in reaction to a wide selection of intra-model and post-model debiasing techniques, each targeting different parts of the models.

\subsection{Corpora}
\label{sec:corpora-main}
Debiasing strategies, especially pre-model ones, largely entail altering the pre-training corpus; thus, we must demonstrate that the BabyLM corpus can support the emergence of the same bias dynamics observed in BERT. We therefore examine corpus aspects known to induce LM biases. This section reports the most critical results; the remainder is listed in Appendix \ref{sec:corpora}.

In terms of the corpora, we utilise the 2023 BabyLM challenge corpus, used to train LTG-BERT, and closely replicate the unavailable BERT corpus from a Wikipedia dump \citep{broad_wiki_2022} and a BookCorpusOpen re-crawl \citep{diliello_bookcorpusopen_2022, bandy2021addressing}, preserving the two-corpora token ratio and the original $\sim$3B-token size.

With the corpora established, we examine differences in their topical coverage \citep{chang-etal-2019-bias, zhao-etal-2019-gender}. For each topic (e.g., race, gender), we compute the percentage of each corpus formed by keywords related to its subcategories (e.g., male, female) \citep{meade2021empirical}. Gender is the most prominent category. Table \ref{tab:gender} shows that the male gender is more represented than the female gender across corpora, with the resulting models also biased in a male-centric direction. This suggests that they should react to gender-focused debiasing in the same way.

\begin{table}[h!]
\centering
\resizebox{\columnwidth}{!}{
\begin{tabular}{l|c|c}
\toprule
\textbf{Category} & \textbf{BabyLM} & \textbf{BERT} \\
\midrule
Male   & \barcell{2.07}{RoyalBlue!70}     & \barcell{1.83}{RoyalBlue!70} \\
Female & \barcell{1.03}{Orange!85!black}  & \barcell{1.30}{Orange!85!black} \\
\bottomrule
\end{tabular}}
\caption{\label{tab:gender}Gender representation}
\vspace{-0.5em}
\end{table}

Across other bias categories, most trends are shared between the corpora. In ethnicity-term frequency, Caucasian terms are consistently most over-represented, Black-related terms second, and Asian strongly last. In religion, Christian terms dominate, while Jewish terms are barely represented in both. LGBTQ+ topics appear equally. The differences lie in the BERT corpus representing Black and Muslim terms much more substantially. Overall, despite BERT containing higher frequencies of biased terms across categories, the BabyLM corpus largely preserves similar topic ratios, indicating it ought to support debiasing techniques.
\begin{table}[h!]
\centering
\resizebox{\columnwidth}{!}{%
\begin{tabular}{l|c|c}
\toprule
\textbf{Category} & \textbf{BabyLM} & \textbf{BERT} \\
\midrule
Black     & \barcells{0.035}{RoyalBlue!70}     & \barcells{0.062}{Orange!85!black} \\
Caucasian & \barcells{0.049}{RoyalBlue!70}     & \barcells{0.067}{Orange!85!black} \\
Asian     & \barcells{0.014}{RoyalBlue!70}     & \barcells{0.034}{Orange!85!black} \\ \hline \hline
Jewish    & \barcells{0.008}{RoyalBlue!70}     & \barcells{0.007}{Orange!85!black} \\
Christian & \barcells{0.034}{RoyalBlue!70}     & \barcells{0.065}{Orange!85!black} \\
Muslim    & \barcells{0.007}{RoyalBlue!70}     & \barcells{0.028}{Orange!85!black} \\ \hline \hline
LGBTQ+     & \barcells{0.017}{RoyalBlue!70}     & \barcells{0.022}{Orange!85!black} \\
\bottomrule
\end{tabular}}
\caption{Demographic/religion mentions}
\vspace{-0.5em}

\end{table}
\\
Finally, as toxicity and hate-speech are linked to increased biases \citep{workshop2022bloom}, we used established models \citep{hanu_2020_7925667, antypas-camacho-collados-2023-robust} to label their presence in each corpus. Table \ref{tab:tox} shows that the BabyLM corpus is more toxic and hateful, even containing blatantly racist and sexualised terms, contrary to its child-aligned disposition.
\begin{table}[h!]
\centering
\resizebox{\columnwidth}{!}{%
\begin{tabular}{l|c|c}
\toprule
\textbf{Corpus} & \textbf{Toxic} & \textbf{Hate-speech} \\
\midrule
BabyLM      & \barcelltox{3.12}{RoyalBlue!70}     & \barcellhate{0.55}{RoyalBlue!70} \\
BERT & \barcelltox{0.95}{Orange!85!black}  & \barcellhate{0.34}{Orange!85!black} \\
\bottomrule
\end{tabular}}
\caption{\label{tab:tox}Toxicity and hate-speech sentence rates}
\vspace{-1em}
\end{table}

Overall, compared to the BERT corpus, BabyLM exhibits similar topic frequencies and a higher presence of toxic and hateful sentences, enabling the same bias and debiasing dynamics. Consequently, the BabyLM corpus appears capable of supporting both bias acquisition and debiasing similar to those observed with BERT.

\subsection{Debiasing behaviour}
Building on the finding that the BabyLM corpus aligns with the BERT corpus across all bias-relevant metrics, we test whether the two model classes debias comparably. To analyse behavioural overlaps, we use a broad set of intra- and post-model debiasing techniques, each using different mechanisms and targeting distinct model components. In each test, we compare how debiasing shifts LTG-BERT’s and LTG-Baseline’s composite bias and performance relative to BERT.

Starting with \underline{post-model debiasing}, we apply two debiasing methods, \textbf{Sent-Debias} and \textbf{INLP}, both targeting trained models with already frozen encoders. Sent-Debias learns a gender subspace from text and subtracts it from the final hidden representations. INLP trains linear classifiers for protected attributes and iteratively projects out the directions that make those attributes linearly separable. The implementation of both approaches uses a 2.5M-word Wikipedia dump for their debiasing signals \cite{meade2021empirical}.

\begin{figure}
    \centering
    \includegraphics[width=1\linewidth]{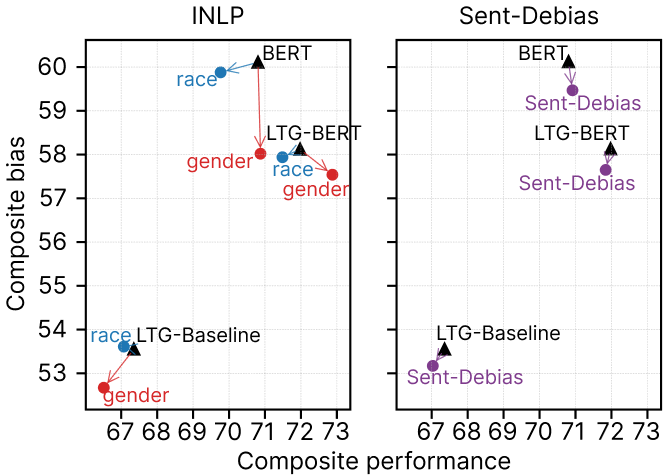}
    \caption{Bias and performance changes caused by post-model debiasing strategies}
    \label{fig:postdeb}
    \vspace{-1em}
\end{figure}

Looking at the impact of debiasing in Figure \ref{fig:postdeb}, we see that Sent-Debias consistently reduces composite gender bias across all models, although the performance impact varies. INLP remains consistent across architectures regarding its effects on bias. Gender-focused INLP lowers overall bias, while race-focused INLP does not reduce composite bias and also harms accuracy. The gender-focused INLP's performance effects align with model fit: the over-fitted LTG-BERT benefits from it as a form of regularisation, achieving SOTA performance (Appendix \ref{sec:INLP-reduction}), the more under-fitted LTG-Baseline loses useful information, and full-data BERT incurs no severe penalties. With this behavioural nuance, we conclude that the methods' impact on bias is consistent across models.
\\
Next, we probe \underline{intra-model debiasing} with four strategies. \textbf{CDA} balances a 10M-word Wikipedia dump by duplicating sentences containing gendered or racial terms and swapping them \cite{meade2021empirical}. \textbf{CDS} uses the gender-balanced corpus, substituting each gender mention with the opposite to create anti-stereotype contexts \citep{webster-etal-2018-mind, bartl2020unmasking}. A \textbf{debiasing-loss} setup trains on the Gender Pronoun Resolution task with Stereotype Neutralisation and Elastic Weight Consolidation \cite{zhao-etal-2018-gender, Park23Never}. A \textbf{dropout} variant increases dropout during continued pre-training on the same Wikipedia dataset.

\begin{figure}
    \centering
    \includegraphics[width=1\linewidth]{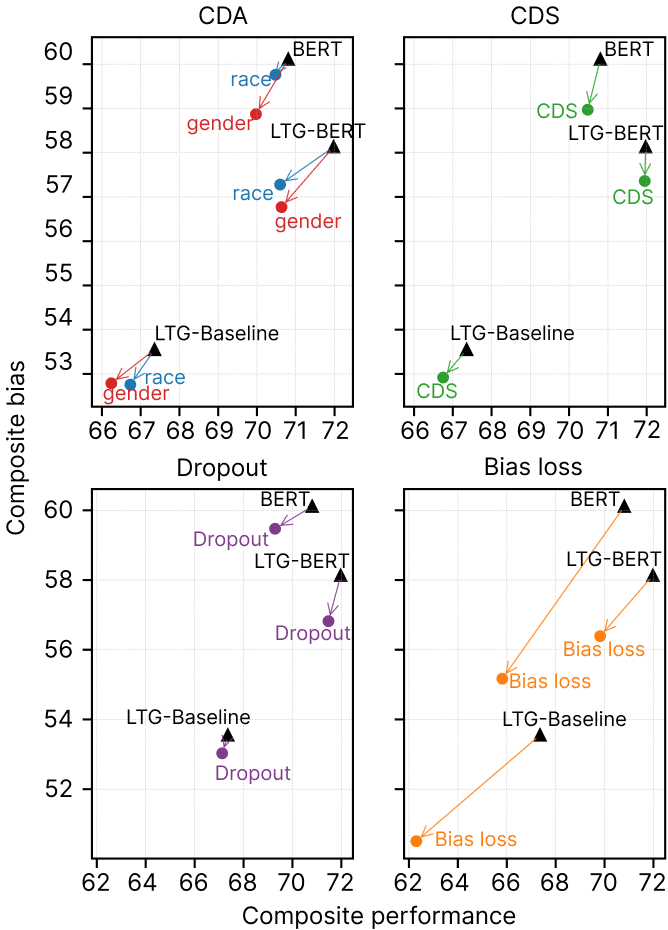}
    \caption{Bias and performance changes caused by intra-model debiasing strategies}
    \label{fig:itradeb}
\end{figure}

Across methods, Figure~\ref{fig:itradeb} shows that the models shift in the same direction, differing mainly in magnitude. CDA on gender and race reduces composite bias across models, with stronger gender effects and similar performance-loss trends. CDS again yields near-identical bias reduction across models with modest, method-consistent accuracy costs. Debiasing loss induces the largest accuracy drop and bias decrease, with LTG-Baseline tracking BERT most closely. Dropout yields weaker debiasing while preserving the trends; some noise in the performance–bias-loss ratio is expected since BERT lacks the extra normalisation of LTG models \citep{shleifer2021normformer}, leading to over-confident logits that drift under perturbation \citep{pmlr-v48-gal16, kong-etal-2020-calibrated}.

These convergent effects show that BabyLMs simulate the behaviour of post- and intra-model debiasing techniques on BERT. We quantify alignment by pairing each model’s performance and bias shifts per method and running canonical correlation analysis \citep{da6385d2-9c65-3860-bbcd-b821fdff69ff}. The results in Table \ref{tab:bias-shift} show that LTG-Baseline is the most faithful proxy of BERT’s debiasing behaviour, whereas LTG-BERT’s over-fitting likely dampens alignment.
\begin{table}[h!]
\centering
\begin{tabular}{l|c}
\toprule
\textbf{Model pair} & \textbf{Correlation(ρ$_1$)} \\
\midrule
BERT$\leftrightarrow$LTG-BERT        & 0.796 \\
\textbf{BERT$\leftrightarrow$LTG-Baseline}   & \textbf{0.981} \\
LTG-BERT$\leftrightarrow$LTG-Baseline & 0.772 \\
\bottomrule
\end{tabular}
\caption{\label{tab:bias-shift}First canonical correlation (ρ$_1$) between performance and bias shifts across all debiasing methods}
\vspace{-1em}
\end{table}

\section{Pre-Model Debiasing Experiments}
In the previous section, we showed that standard LMs and BabyLMs share debiasing dynamics, enabling us to estimate a debiasing method’s impact on a standard LM by applying it to a BabyLM. We can now utilise this to run pre-model debiasing experiments on LTG-Baseline instead of BERT, reducing the cost from 500 GPU-hours per experiment to just over 30.

Throughout this section, we reinforce that BabyLM mimics BERT’s reported behaviour and show how this benefits future pre-model debiasing investigations. All experiments use the LTG-Baseline architecture and alter only the training corpus (pre-training setup details in the Appendix~\ref{sec:hyper}).

We establish a baseline by training the LTG model on the original BabyLM corpus, tracking performance and bias over time. During this baseline training, the model picked up bias early, which then stabilised at a steady level, while performance improved more gradually as it learned richer linguistic structure. The quick uptake of bias suggests that the biases likely stem from the topical imbalances and stereotypes, which most pre-model debiasing techniques target. 

\subsection{CDA Pre-model Debiasing}
\begin{figure}[h!]
    \centering
    \includegraphics[width=1\linewidth]{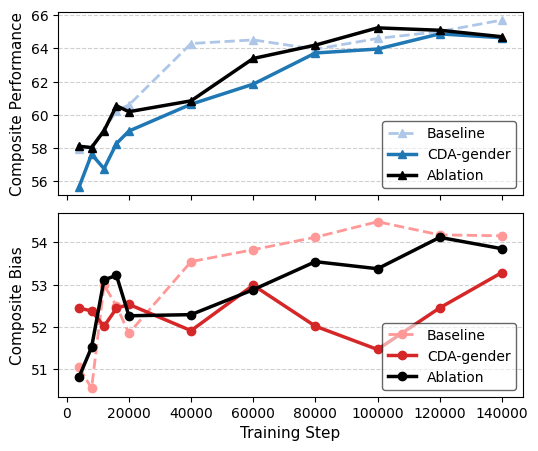}
    \caption{Bias and performance metrics evaluation from pre-training on the CDA and CDA-ablation corpora}
    \label{fig:cdaplot}
    \vspace{-0.75em}
\end{figure}

\begin{figure}[h!]
    \centering
    \includegraphics[width=1\linewidth]{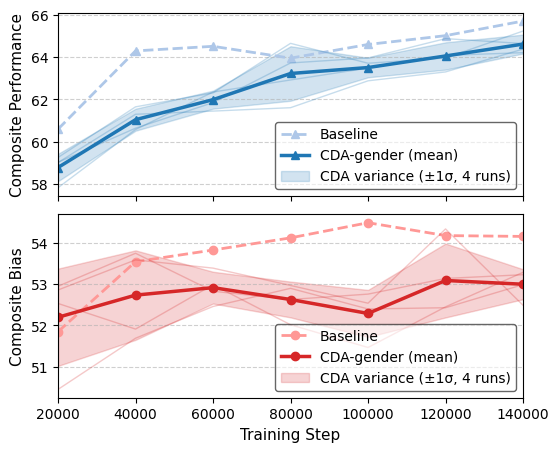}
    \caption{Bias and performance variance during CDA pre-training across random seeds \{1, 10, 50, 100\}}
    \label{fig:cdarob}
    \vspace{-1.5em}
\end{figure}

As a first experiment, we apply pre-model CDA: for every sentence containing a gendered term, we append a flipped-gender counterpart, increasing corpus size by $\sim$59\% and creating a gender-balanced corpus. To check whether any pre-training effects come from balancing rather than simple duplication, we run an ablation that duplicates an equal number of random BabyLM corpus sentences. Figure \ref{fig:cdaplot} shows that the CDA initially slows the model’s grasp of some linguistic concepts but, with longer training, reaches near-baseline performance while clearly curbing bias by preventing its observed steady growth. In contrast, the ablation results in the same performance drop while producing only a small bias reduction. Overall, this validates CDA as a sound debiasing strategy and highlights BabyLMs as a cost-effective platform for such controlled experiments.
\\
To assess the robustness of the BabyLM-based sandbox, we repeat CDA pre-training with multiple random seeds while keeping everything fixed. Figure~\ref{fig:cdarob} shows consistent bias and performance trajectories across runs, with only minor variation and no change to the overall conclusions.
\subsection{Toxicity Removal}
Next, we examine the suggested but unproven claim that corpus toxicity directly drives model bias within LLMs. With 3.39\% of BabyLM sentences being toxic or hateful, we seek to push toxicity to 0\% via two interventions.

Firstly, we utilise an LLM (Llama-3.3-70B) to rewrite toxic sentences while preserving text meaning, discarding only 0.42\% of sentences that could not be detoxified (implementation detailed in Appendix \ref{sec:LLM}). This yielded a slight performance gain and a surprisingly small bias drop, possibly because it imported the LLM’s style and biases into the corpus. Second, we dropped all toxic sentences, which significantly reduced bias and slightly harmed performance.

\begin{figure}[h!]
    \centering
    \includegraphics[width=1\linewidth]{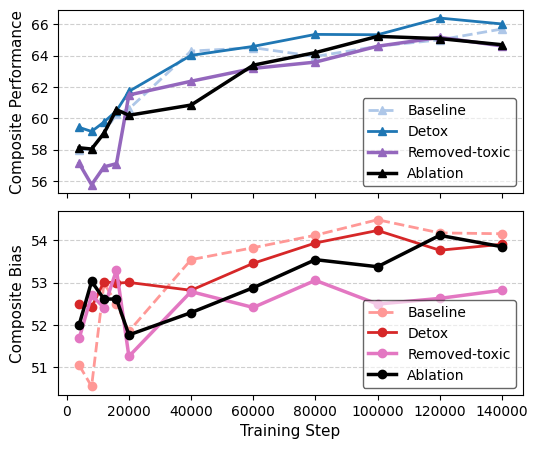}
    \caption{Evolution of bias and  performance during pre-training under three corpus strategies: LLM detoxification, removing toxicity, and toxicity-removal ablation}
    \label{fig:detox}
    \vspace{-2em}
\end{figure}

Last, our ablation test removing an equal number of non-toxic sentences matched the performance drop but failed to match the bias decrease, showing that eliminating toxicity itself, rather than corpus shrinkage, drives the debiasing. Figure~\ref{fig:detox} summarises these results, establishing a clear link between toxicity and standard bias and highlighting the advantage of our BabyLM-based approach that allows us to identify such behaviour.

\section{Perturbation Augmentation}
Finally, we evaluate perturbation augmentation \cite{qian-etal-2022-perturbation}, which uses a perturber LM to rewrite the corpus by randomly swapping demographic references (race and gender), thereby equalising topic–demographic co-occurrence (implementation discussed in Appendix~\ref{sec:perturb-implement}).

\begin{figure}[h!]
    \centering
    \includegraphics[width=1\linewidth]{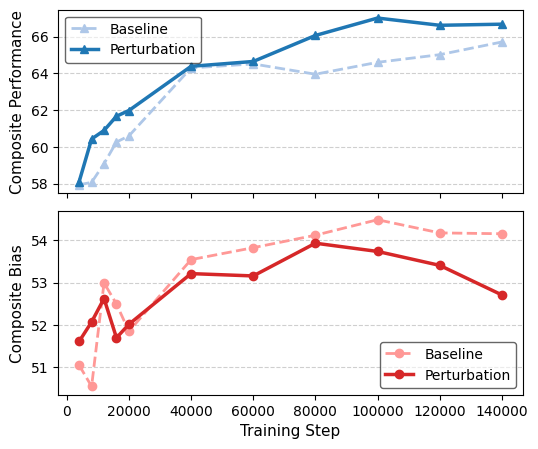}
    \caption{Bias and performance metrics evolution from pre-training on the perturbed corpus}
    \label{fig:perturb}
        \vspace{-0.5em}
\end{figure}

In the original study, pre-training RoBERTa on perturbed data reduced bias substantially with a slight performance gain. Using $\sim$800$\times$ fewer GPU-hours, we closely reproduce these trends (Figure~\ref{fig:perturb}). The perturbation outperforms CDA in debiasing, likely by introducing greater lexical and syntactic variety and by covering more attributes. Moreover, it boosts performance, with the small gains possibly reflecting improved word order in lower-quality sentences caused by perturbation. Overall, even with this more complex debiasing strategy, BabyLM closely mirrors behaviour reported in far more resource-intensive experiments.
\\
In conclusion, LTG-Baseline successfully reproduced the expected behaviours across all tested pre-model debiasing tasks. Moreover, this setup let us both replicate the original experiments and introduce ablations that examined key components of several debiasing strategies. Consequently, under a strict budget, we were able to conduct experiments that had not previously been attempted.

Thus, we show that BabyLMs, and possibly other similarly positioned models, can be very versatile and representative tools for ascertaining the effectiveness of debiasing methods, greatly lowering both the cost and the time required to conduct the exploration-stage of debiasing experiments.

\section{Findings}
With our analysis complete, we note several key trends that directly support the use of low-cost LMs to democratise research on pre-model debiasing.

\paragraph{Finding 1: BabyLM corpora and architectures share the same tendencies regarding bias–performance dynamics as standard LMs.}
Despite far smaller training sets and adapted architectures, BabyLMs acquire linguistic knowledge and biases along the same trajectories as standard BERT-style LMs. Across models, \emph{composite performance} and \emph{composite bias} are strongly and similarly correlated (Table~\ref{tab:cor}). Corpus analysis shows that the BabyLM data, although grammatically simpler, contains the same well-known bias instigators (gender imbalance, topical skew, toxicity, etc.). In particular, male terms are over-represented, Christian terms dominate religious mentions, and toxicity and hate-speech rates are higher than in our reconstructed BERT corpus. Consequently, the SOTA BabyLMs match both BERT's performance and bias, while lower-resource BabyLMs preserve the same bias–ability correlation. This makes such models viable low-cost sandboxes for bias studies.

\paragraph{Finding 2: The debiasing behaviour of BabyLMs strongly correlates with that of standard LMs.}
Building on the first finding, we compared post-model and intra-model debiasing across BERT, LTG-BERT, and LTG-Baseline to test whether BabyLMs share the same debiasing behaviour as standard LMs. Across all debiasing methods, BabyLMs exhibited decreases in bias closely resembling BERT. Greater variance, on the other hand, was observed in performance changes, with post-model approaches being especially sensitive to architectural and pre-training differences between models. Nevertheless, canonical correlation analysis of bias and performance shifts across all methods showed near-perfect alignment between BERT and LTG-Baseline and strong alignment with LTG-BERT. Thus, we show that even a BabyLM trained on 30 GPU-hours closely proxies debiasing dynamics observed in full-scale models.

\paragraph{Finding 3: BabyLMs enable informative and far more cost-effective pre-model debiasing research.}
Finally, we demonstrate the utility of our proposed approach by running seven pre-training interventions on LTG-Baseline, reproducing established results while enabling new experiments and targeted ablations. We use these interventions both to validate our method and to illustrate how it can be used to benefit pre-model debiasing research. We replicate prior findings that CDA reduces bias but can slow learning, and that perturbation augmentation yields larger bias reductions without degrading performance. Beyond replication, we directly linked corpus toxicity to downstream bias, making this the first study in a pre-trained model setting to not only imply but directly test the effect of removing toxic sentences on the resulting bias. With additional ablation experiments, we then isolated specific bias causes (gender imbalance, toxicity), showing that debiasing helps primarily by addressing these instigators rather than incidental corpus restructuring. Crucially, the same bias–performance trends persist across runs with different random seeds, indicating that the observed effects are not a one-off consequence of training stochasticity and supporting BabyLMs as a stable, cost-effective sandbox for pre-model debiasing research under limited compute.

\section{Conclusion}
This work raised the issue that most LM debiasing research focuses only on pre-trained models with already-formed biased circuits. We argued that, to develop effective debiasing strategies, we must first understand how bias emerges in LMs and devise approaches that prevent its original formation. Such research is rare due to its prohibitive cost. To fix this, we proposed investigating debiasing dynamics in well-performing, low-resource, data-efficient models, such as BabyLMs.

Our experiments showed that BabyLMs use sufficient data and acquire intrinsic biases comparably to standard LMs when matched for performance. Their debiasing behaviour likewise mirrors other LMs, indicating that BabyLMs’ democratised pre-training setup does not disqualify them. Thus, we moved from pre-training costs of 500 GPU-hours to 30 GPU-hours using a BabyLM, creating a pathway for affordable debiasing research. With this, we reproduced previously reported results and added findings that solidified toxicity and bias imbalance as major contributing factors to LM bias. 

Overall, our hope is that this research encourages the use of low-cost LMs that enable the exploration and negation of bias formation, allowing researchers to identify promising methods before committing to costly large-scale experiments and to move the entire LM debiasing field forward.

\section{Limitations}
One limitation of this study is the set of metrics it was able to use. BabyLMs limit us to more simplistic bias and performance-evaluation frameworks, since they do not offer sufficiently developed language and world understanding to support advanced extrinsic evaluations. Furthermore, due to the large amount of evaluation throughout the entire paper, frameworks like SuperGLUE, which take 2–3 hours to run on our hardware, are infeasible. In addition, all the metrics we used are English-specific. This is important to note because, even though non-English versions of these benchmarks exist \cite{neveol-etal-2022-french, ozturk2023different}, they are very limited and still do not cover low-resource languages, which is a major obstacle for LM bias research.

Secondly, it is well discussed that bias evaluations represent only a subset of the larger issue, and there might be biases where the different models’ behaviours actively differ but cannot be observed. Thus, we recommend using the composite bias metric to identify overall trends, but there may be a need to investigate or propose more specialised tests when tracing specific types of biases.

Related to this, while BabyLMs show promising alignment, there will be tasks that they cannot replicate. As noted in the paper, we propose them as a tool for \textit{exploration}, helping to identify promising debiasing strategies. When these strategies are identified, they still need to be validated on larger models. Nevertheless, BabyLMs still allow us to skip the costly experimental phase.

Finally, it should be noted that BabyLMs were used as a promising and easily available set of architectures that displayed appropriate properties for the task at hand. We still encourage efforts to explore different corpora and architectures. CLMs, especially in the form of LLMs, cannot be replaced by MLMs. As such, future work should propose a model that democratises debiasing research for CLMs. Likewise, there might be corpora even better suited for testing debiasing methods than the BabyLM one. This study simply shows that the promising results obtained with the LTG-Baseline and the BabyLM corpus provide strong evidence that this path towards the democratisation of debiasing research is possible and promising.

\section{Ethical Considerations}
This work does not involve human-subject data or sensitive personal information. Nevertheless, any discussion of bias and toxicity in language models must acknowledge that our evaluation is incomplete and grounded in prior definitions of bias. Accordingly, we do not claim that any debiasing approach is definitive, nor that even the best-performing strategies remove all forms of bias.

As noted, implementations of our framework should not proceed under the assumption that it is sufficient on its own. In particular, when debiasing a publicly available model, we must ensure that biases are also evaluated in the model itself rather than purely relying on cheaper proxies.

\section*{Acknowledgements}
This paper reports on work supported by Cambridge University Press \& Assessment. We thank colleagues in the ALTA Institute for their support and feedback.

\bibliography{custom}

\appendix

\section{Evaluated Models}
\label{sec:eval-models}
In order to run our experiment, we must analyse a sufficiently large set of BabyLMs and standard LMs, so that we can observe meaningful trends.

Regarding BabyLMs, we collect every notable BabyLM that is available online with executable code, utilising our overview in Section~\ref{sec:BabyLMs} \citep{samuel2023trained}. It should be noted that some models require altered data input pipelines, making them impractical. Likewise, several promising papers never released their models. To ensure comparability, we take only models from the \texttt{strict} track. The final list is shown in Table~\ref{tab:BabyLM-Models}.

\begin{table*}[h!]
\centering
\begin{tabular}{l|p{7cm}}
\toprule
\textbf{Model key} & \textbf{Hugging Face ID}\\
\midrule
BabyLM2024 & \texttt{jdebene/BabyLM2024} \\
elc-bert & \texttt{lgcharpe/ELC\_BERT\_baby\_100M}\\
cambridge-climb & \nolinkurl{cambridge-climb/baseline-roberta_pre_layer_norm-model} \\
gpt-bert-babylm & \texttt{ltg/gpt-bert-babylm-base} \\
ltg-bert-babylm & \texttt{ltg/ltg-bert-babylm} \\
ltgbert-100m-2024 & \texttt{babylm/ltgbert-100m-2024} \\
roberta-base-strict-2023 & \texttt{babylm/roberta-base-strict-2023} \\
babyllama-100m-2024 & \texttt{babylm/babyllama-100m-2024} \\
baby-llama-2-345m & \texttt{JLTastet/baby-llama-2-345m} \\
\bottomrule
\end{tabular}
\caption{\label{tab:BabyLM-Models}Examined BabyLMs with their Hugging Face IDs}
\end{table*}

For standard LMs, we also need to observe how bias scales with performance. Testing just a few primary models would not establish a trend. To resolve this, we examine the popular architectures used to create BabyLMs together with other variants of these architectures trained on different corpora (e.g.\ legal documents \citep{chalkidis-etal-2020-legal}, Twitter \citep{barbieri-etal-2020-tweeteval}, scientific papers \citep{beltagy-etal-2019-scibert}). All these models are listed in Table~\ref{tab:Standard-Models}.
\begin{table*}[h!]
\centering
\begin{tabular}{l|p{7cm}}
\toprule
\textbf{Model key} & \textbf{Hugging Face ID}\\
\midrule
BiomedBERT-abstract &
{%
  \def\UrlBreaks{\do\/\do\-\do\_}\Urlmuskip=0mu plus 1mu
  \nolinkurl{microsoft/BiomedNLP-BiomedBERT-base-uncased-abstract}%
} \\
BiomedBERT-abstract-fulltext &
{%
  \def\UrlBreaks{\do\/\do\-\do\_}\Urlmuskip=0mu plus 1mu
  \nolinkurl{microsoft/BiomedNLP-BiomedBERT-base-uncased-abstract-fulltext}%
} \\
DialoGPT-large & \nolinkurl{microsoft/DialoGPT-large} \\
DialoGPT-medium & \nolinkurl{microsoft/DialoGPT-medium} \\
DialoGPT-small & \nolinkurl{microsoft/DialoGPT-small} \\
bert-base-cased & \nolinkurl{bert-base-cased} \\
bert-base-uncased & \nolinkurl{bert-base-uncased} \\
bert-for-patents & \nolinkurl{anferico/bert-for-patents} \\
BNC bert & \nolinkurl{ltg/ltg-bert-bnc} \\
finbert & \nolinkurl{yiyanghkust/finbert-pretrain} \\
gpt2-medium & \nolinkurl{openai-community/gpt2-medium} \\
gpt2-xl & \nolinkurl{openai-community/gpt2-xl} \\
legal-bert & \nolinkurl{nlpaueb/legal-bert-base-uncased} \\
roberta-base & \nolinkurl{FacebookAI/roberta-base} \\
scibert & \nolinkurl{allenai/scibert_scivocab_uncased} \\
twitter-roberta-base & \nolinkurl{cardiffnlp/twitter-roberta-base} \\
\bottomrule
\end{tabular}
\caption{\label{tab:Standard-Models}Examined standard models with their Hugging Face IDs}
\end{table*}

\section{Model Alignment}
\label{sec:align-models}
\begin{figure}[h!]
    \centering
    \includegraphics[width=1\linewidth]{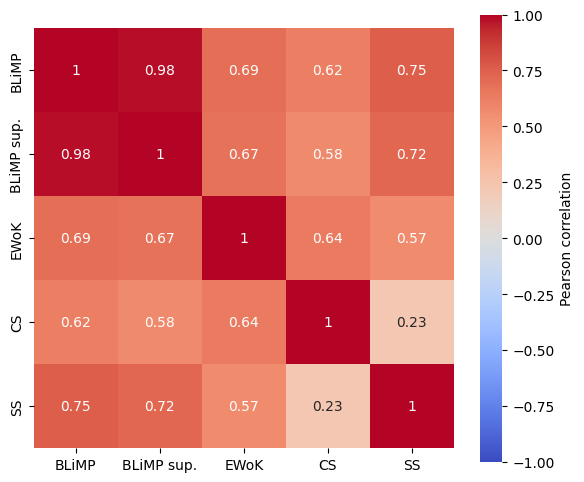}
    \caption{\label{fig:cor-baby}Pearson correlations (BabyLM models, $N=9$)}
\end{figure}

\begin{figure}[h!]
    \centering
    \includegraphics[width=1\linewidth]{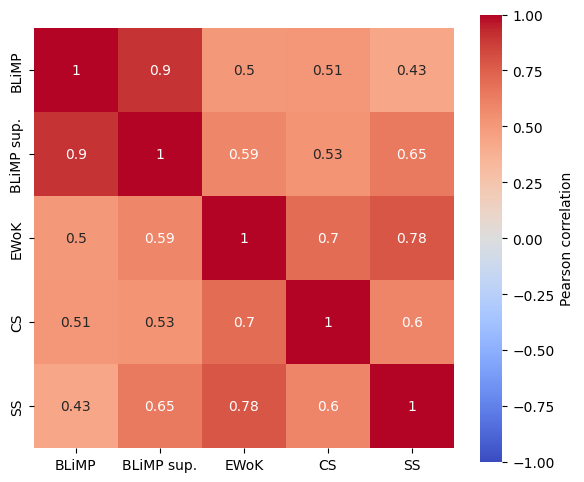}
    \caption{\label{fig:cor-stand}Pearson correlations (standard models, $N=16$)}
\end{figure}

Running all models through our pipelines, we obtain Pearson correlations between the bias metrics and the performance metrics. Figures~\ref{fig:cor-baby} and~\ref{fig:cor-stand} show the correlation results for BabyLMs and standard LMs respectively. In both groups, performance and bias metrics are strongly positively correlated, while the two bias scores have a positive but more limited correlation with each other. BabyLMs show especially noisy behaviour since some of them have very low word knowledge needed to display biases. This leads us to establish the composite metrics, which more completely capture the overall behaviour.

\section{Selecting Candidate BabyLMs}
\label{sec:selecting-models}
\begin{figure*}[h!]
    \centering
    \includegraphics[width=1\linewidth]{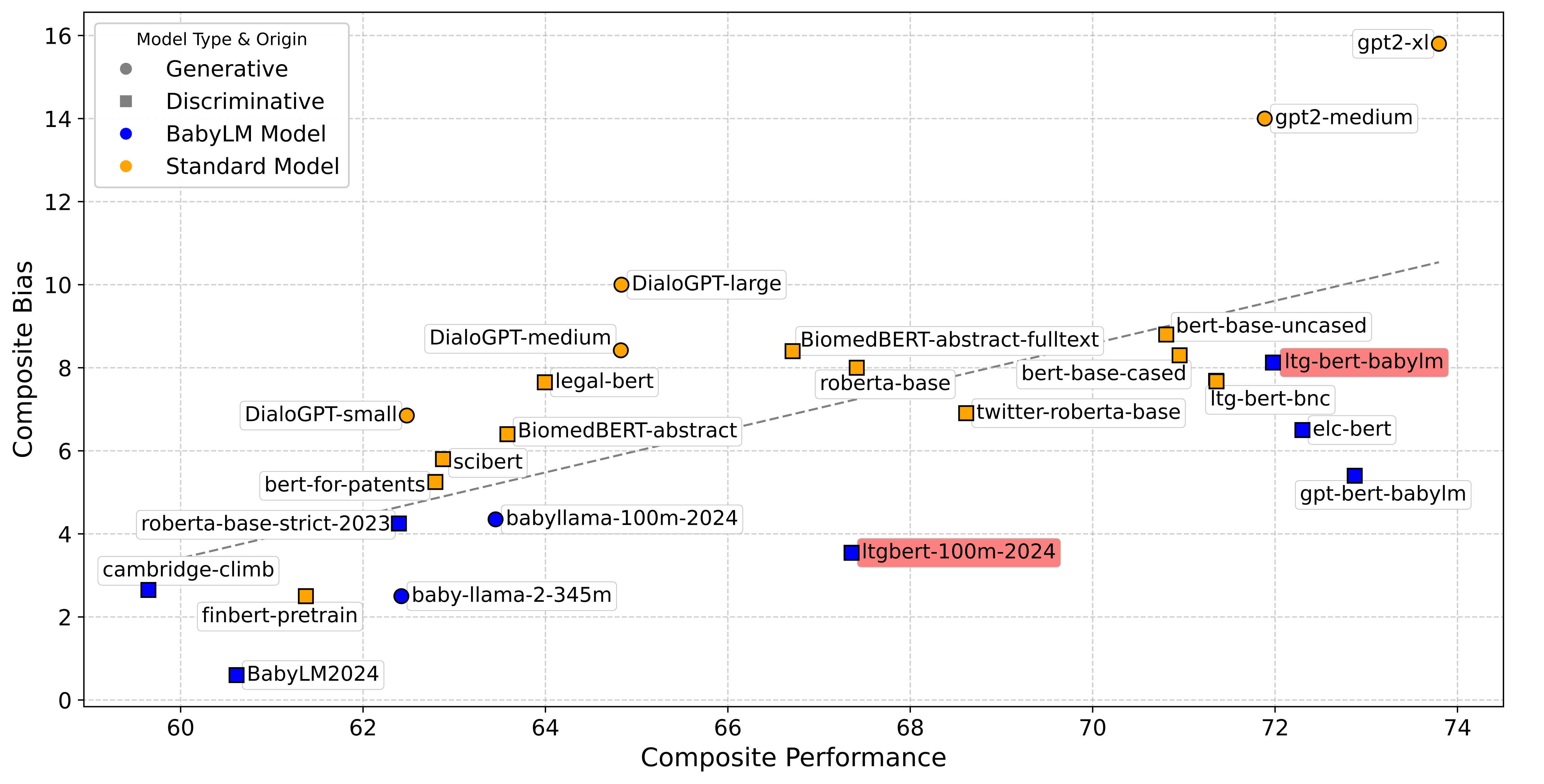}
    \caption{Average performance correlated with average bias for all evaluated LMs, with the average trend being shown and the candidate models being highlighted}
    \label{fig:bias_vs_avg}
\end{figure*}

Furthermore, we need to select the most informative and relevant BabyLMs, which we continue using in our research. To do so, we are interested in models that are as close to the original BERT as possible in both behaviour and architecture, making them suitable for replicating BERT’s behaviour. To this end, we examine Figure~\ref{fig:bias_vs_avg}, which visualises the results obtained in the previous section, and we highlight two promising models. 

Of all the models considered, ltg-bert-babylm is the most closely aligned with BERT’s architecture, as indicated by the description in Section~\ref{sec:BabyLMs} \citep{samuel2023trained}. It has no modifications that would make it more complicated or less predictable to work with.

Nevertheless, our ultimate goal is to advance democratisation research, and ltg-bert-babylm, together with the other high-performing models, was pre-trained for roughly the same number of GPU-hours as the original BERT. Thus, although ltg-bert-babylm is ideal for studying how debiasing operates in the Babylm-BERT relation, it is less suitable for democratisation research.

\begin{figure}
    \centering
    \includegraphics[width=1\linewidth]{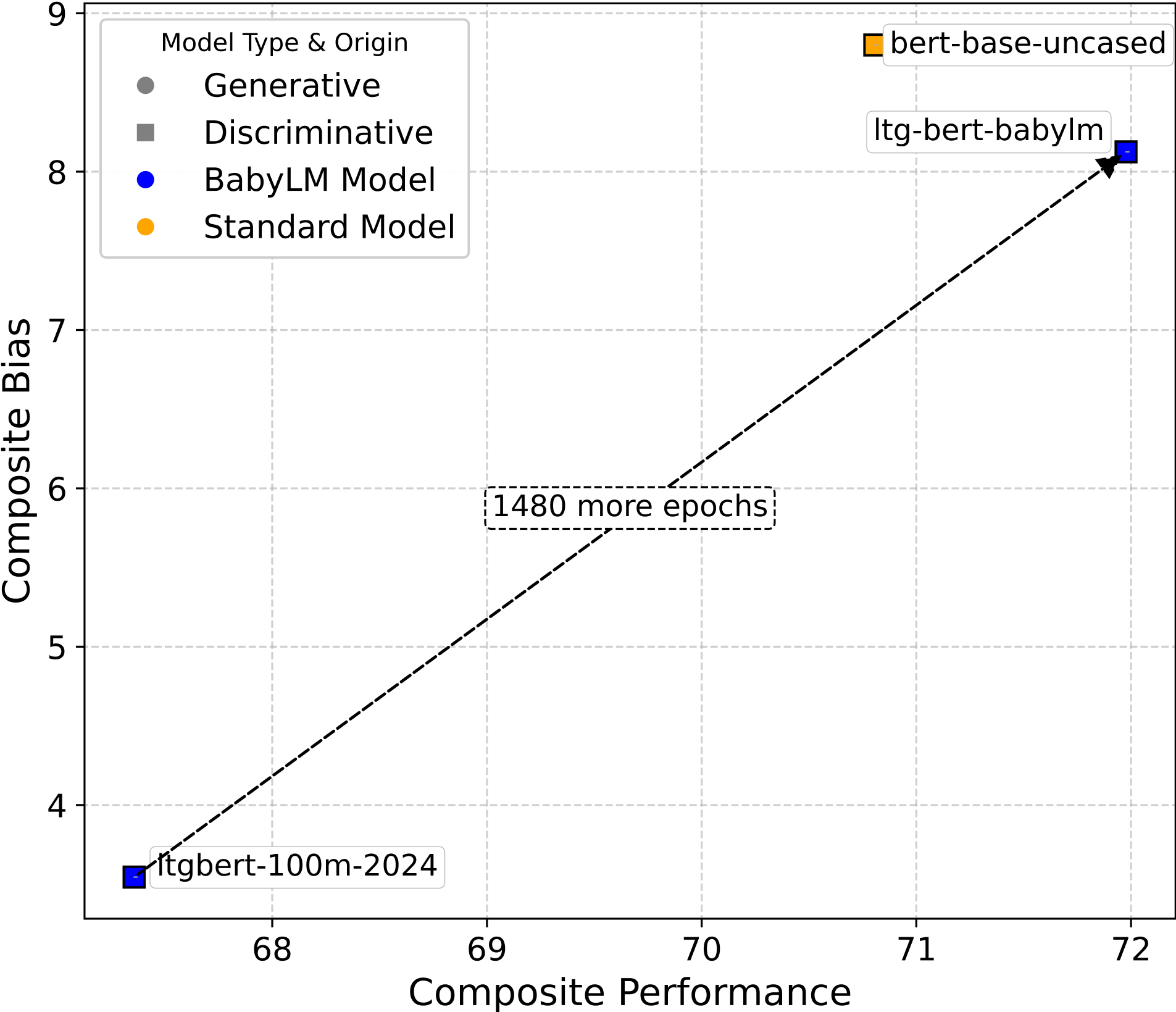}
    \caption{Performance–bias trade-off for ltg-bert-babylm and its low-cost baseline}
    \label{fig:ltg-change}
\end{figure}

This limitation was recognised even by the 2024 BabyLM organisers, who released ltgbert-100m-2024 \citep{hu-etal-2024-findings}. This model is a baseline version of the SOTA ltg-bert-babylm. It was trained for only 20 epochs (versus 1,500 epochs for the SOTA model) and required less than 40 GPU-hours. Figure~\ref{fig:ltg-change} shows the contrast between the two versions. Although ltgbert-100m-2024 is notably further from BERT in performance, it still achieves reasonable performance scores and exhibits measurable bias, making it a good baseline for our experiments.
\section{Corpora}
This section details all other experiments conducted with the aim of exploring and comparing the corpora of BabyLMs and standard LMs.
\label{sec:corpora}
\subsection{Structural and Syntactic Metrics}
We start the analysis by examining the basic linguistic metrics present in the corpora, trying to understand the composition and complexity of each dataset. This might help us understand whether the BabyLM corpus is sufficiently coherent to form a full range of biases.

To do this, we leverage the TextDescriptives library by \citet{hansen2023textdescriptives}, which allows us to derive more than sixty syntactic and discourse-level statistics. Because all corpora exceed the toolkit’s token limit for text processing, we split the text into chunks of 8192 tokens, retaining large enough samples to assess the more structural metrics while keeping the problem tractable. In the end, the outputs per chunk are averaged into final values representing the entire corpus. This ensures that no corpus is penalised for its size while capturing the structural and syntactic profile needed for analysis.
\subsubsection{Part-of-Speech Profiles}
PoS statistics give us a compact high-level overview, informing us about the average type of the texts present in each corpus. We are especially interested in the frequency of function words (e.g., pronouns, determiners), which organise discourse, versus the frequency of content words (e.g., nouns, verbs), which carry meaning.

\begin{figure*}[h!]
    \centering
    \includegraphics[width=1\linewidth]{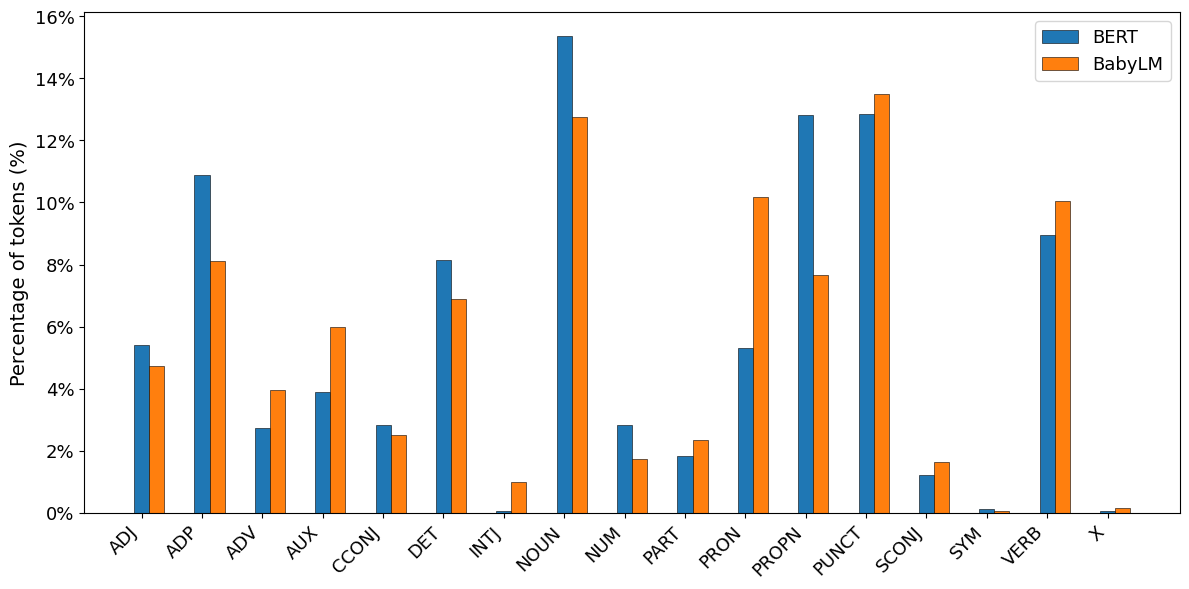}
    \caption{\label{fig:pos-profile}Part-of-Speech distribution across corpora}
\end{figure*}

Figure~\ref{fig:pos-profile} shows us that BabyLM displays a lower frequency of nouns and pronouns, signalling a simpler and more narrative-driven corpus.

\begin{figure}[h!]
    \centering
    \includegraphics[width=1\linewidth]{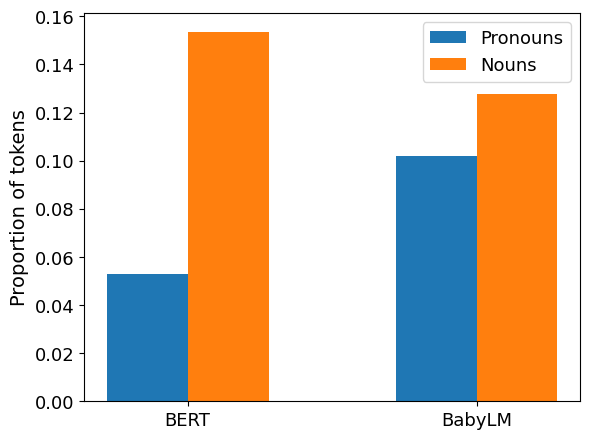}
    \caption{\label{fig:pvn-profile}Pronoun vs.\ noun distribution across corpora}
\end{figure}

Figure~\ref{fig:pvn-profile} serves to further highlight this contrast. BabyLM’s proportion of pronouns to nouns is, unlike that of BERT, almost at parity. Thus, in the BabyLM corpus, more tokens are used to refer to individuals rather than describing unique objects or concepts, again indicating that it is linguistically simpler than the BERT corpus.
\subsubsection{Linguistic Complexity and Readability}
Since it appears that the main difference is text complexity, we investigate the readability metrics.

We choose to use the Flesch–Kincaid Grade Level (FKGL), which is computed from the average sentence length and the average number of syllables per word. Its final value then corresponds to grades in the US school system, ranging between 0 and 18. With the results shown in Table~\ref{tab:corpus_readability_stats}, we can once again see that BERT has the linguistically more advanced texts and BabyLM has the simpler ones. The FKGL places the BabyLM corpus between 5th and 6th grades, which corresponds to students aged 10-12 and thus aligns with what would be expected for BabyLM. 
\begin{table*}[h!]
\centering
\begin{tabular}{l|c|c|c|c}
\toprule
\textbf{Corpus} & \textbf{Avg.\ syllables per word} & \textbf{Avg.\ word length} & \textbf{Avg.\ sentence length} & \textbf{FKGL} \\
\midrule
BabyLM & 1.25 & 4.01 & 15.91 & 5.40 \\
BERT   & 1.38 & 4.57 & 19.84 & 8.43 \\
\bottomrule
\end{tabular}
\caption{Readability and length statistics for the BERT and BabyLM corpora.}
\label{tab:corpus_readability_stats}
\end{table*}

\begin{table}[h!]
\centering
\begin{tabular}{lc}
\toprule
\textbf{Corpus} & \textbf{Type--token ratio} \\
\midrule
BabyLM & 0.33 \\
BERT   & 0.53 \\
\bottomrule
\end{tabular}
\caption{Type--token ratio for the BERT and BabyLM corpora}
\label{tab:ttr}
\end{table}

Unsurprisingly, when investigating the type-token ratio shown in Table~\ref{tab:ttr} to ascertain the richness of the vocabulary, we see that BERT has the more diverse vocabulary. BabyLM is again much simpler. This is likely due to its enforced child-alignment.

\subsubsection{Coherence}
Finally, we use the first-order coherence to examine how tightly each text corpus adheres to a topic. In Table~\ref{tab:coherence}, we see that the corpora score highly, with both being similarly consistent.
\begin{table}[h!]
\centering
\begin{tabular}{l|c}
\toprule
\textbf{Corpus} & \textbf{First-order coherence} \\
\midrule
BabyLM & 0.811 \\
BERT   & 0.802 \\
\bottomrule
\end{tabular}
\caption{First-order coherence scores for the BERT and BabyLM corpora}
\label{tab:coherence}
\end{table}
\\
Overall, the BabyLM corpus is markedly simpler in text complexity yet maintains coherence comparable to BERT. Consequently, if it contains sufficient bias-inducing material, it is likely adequate to impart those biases to the model.

\subsection{Toxicity, Sentiment, and Emotions}
\label{sec:toxic}
Furthermore, we need to understand the tone of the corpora and the context in which they present their different topics. While no exact link has been created, there is some evidence of the fact that displaying information in a negative or toxic light pushes the model to develop stronger biases \citep{workshop2022bloom}. As such, we must examine the presence of toxicity, hate-speech, sentiment, and the overall emotional composition throughout the corpora.

\subsubsection{Emotion Scores}
Starting with emotion detection, we are interested in measuring Ekman’s core emotions: joy, sadness, fear, surprise, anger, and disgust \citep{ekman1999basic}. To this end, we take advantage of the \textit{NRC Emotion Lexicon}, which links individual words to various emotions \citep{mohammad-turney-2010-emotions}, allowing us to infer the overall emotional tone of the corpus. The normalised \textit{emotion score} of a corpus is calculated by dividing the number of words expressing a specific emotion by the total number of eligible words. This \textit{emotion score} is computed for all six emotions across all our corpora.

\begin{figure*}[h!]
    \centering
    \includegraphics[width=1\linewidth]{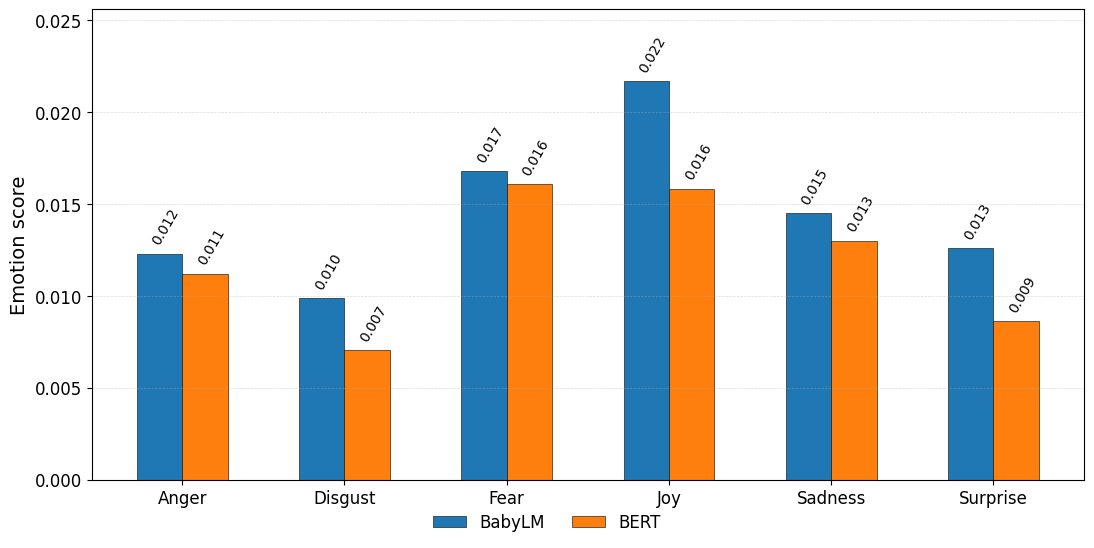}
    \caption{\label{fig:emotions}Distribution of emotions across the corpora using the emotion score obtained by lexical analysis}
\end{figure*}

The results of the lexical analysis can be seen in Figure~\ref{fig:emotions}. While the emotions are mostly balanced across the corpora, BabyLM is consistently more emotional, especially in terms of joy and surprise. However, in both cases, the difference is mild and both are positive emotions, which are not the carriers of bias.

\subsubsection{Sentiment}
Works like the one by \citet{koksal-etal-2023-language} have indicated that sentiment is passed from corpora into the biases of models, forging positive or negative connections that might end up strengthening stereotypes. Because of this, we measure the sentiments in each corpus, both the overall ones and those connected to specific topics.

We use the established RoBERTa-based solution by \citet{barbieri-etal-2020-tweeteval}, which is one of the most popular publicly available sentiment analysis models. With it, we classify each corpus as positive, negative, or neutral. Finally, we compute the \textit{sentiment score} in the same fashion we computed the \textit{emotion score}, only switching from word-based to sentence-based evaluation.

\begin{figure}
    \centering
    \includegraphics[width=1\linewidth]{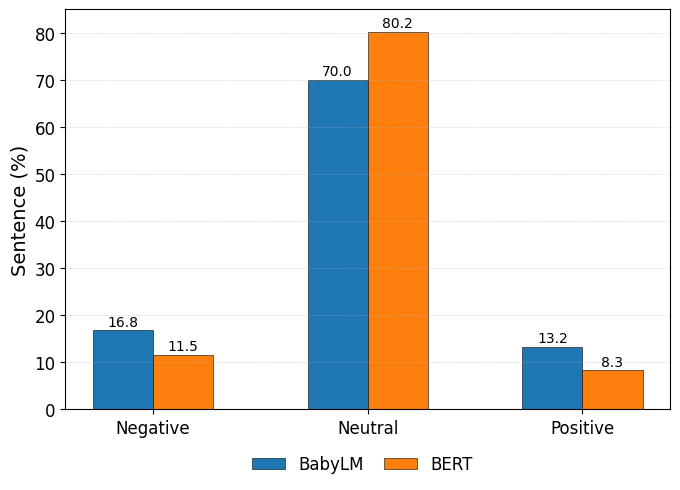}
    \caption{\label{fig:sentiment}Distribution of sentences per sentiment across corpora}
\end{figure}

Figure~\ref{fig:sentiment} shows the overall sentiment profile of the corpora. The BabyLM corpus is consistently less neutral, reinforcing that it is well-positioned to transfer the biases into the model despite its child-alignment.

\begin{figure*}[h!]
    \centering
    \includegraphics[width=1\linewidth]{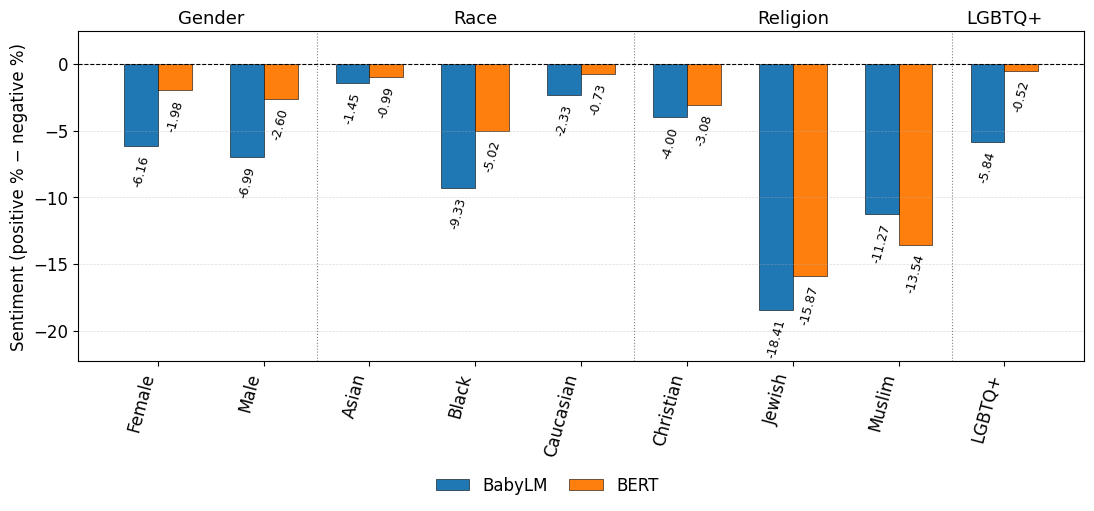}
    \caption{\label{fig:sentiment-topic}Sentiment stances to the selected topics across corpora}
\end{figure*}

Examining the results more closely, Figure~\ref{fig:sentiment-topic} displays the sentiment of sentences containing a bias-inducing word from a specific category for each corpus. To calculate this, we take all eligible sentences per topic and subtract the percentage of negative sentences from the percentage of positive sentences. Here, we can see that the BabyLM corpus keeps being mostly more negative than the BERT corpus, with the exception of the Muslim topic. While it carries more negativity, the ratios between topics are similar, meaning that it retains the same bias orientation. The only true divergence is higher negativity towards LGBTQ+ topics. 

In summary, BabyLM proves to be more emotional and negative than BERT, while retaining the same larger trends, meaning that the corpora are largely compatible. There is no evidence that BabyLM should not be able to teach model biases.

\subsection{Specifications Regarding Toxicity} 
While this topic has already been covered in Section \ref{sec:corpora-main}, we want to provide a validation of the result. Given that we have established that the BabyLM corpus mostly contains simple sentences, the toxicity and hate-speech results in Table \ref{tab:tox} raise the question of whether the toxicity and hate-speech labels are overly sensitive. To investigate this, we sampled 200 such sentences, identifying that the vast majority are indeed toxic, containing slurs or highly aggressive expressions. For illustration, we list some sampled examples in Table~\ref{tab:toxic-sample} (\textbf{Contains explicit offensive statements}). These examples display blatant racist and sexualised terms, showing that, despite its child-alignment, BabyLM contains diverse and toxic texts. 
\begin{table}[h!]
    \centering
        \begin{tabular}{l}
        \hline
        \textbf{Sentence} \\ \hline
        Her earhole ain’t big enough for f***ing! \\
        cause they don’t want to look like morons too. \\ 
        He was a black n****r what? \\  \hline
    \end{tabular}
    \caption{\label{tab:toxic-sample}Sampled examples of toxic sentences from the BabyLM corpus}
\end{table}
\section{INLP Reduces Performance Penalty for Over-Fitted BabyLMs}
\label{sec:INLP-reduction}
The INLP experiment has also revealed an important piece of behaviour from LTG-BERT that concerns the BabyLM challenge itself. As mentioned, INLP helped improve LTG-BERT’s performance, likely by reducing the issues connected to over-fitting on its training corpus.

\begin{figure}[h!]
    \centering
    \includegraphics[width=1\linewidth]{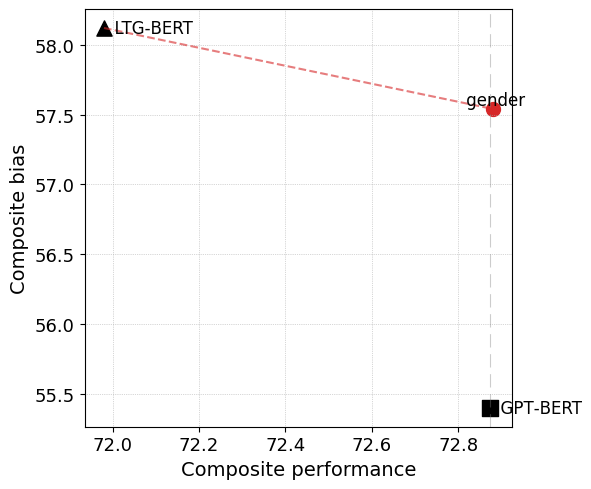}
    \caption{\label{fig:INLP_performance}Performance improvement of INLP-debiased LTG-BERT compared with the GPT-BERT BabyLM}
\end{figure}

This is notable because LTG-BERT is the best-performing BabyLM-class MLM, with the SOTA BabyLM being GPT-BERT, which was trained with the same training data-to-epoch ratio. Thus, if INLP debiasing makes LTG-BERT surpass the SOTA model on our performance tasks, as shown in Figure~\ref{fig:INLP_performance}, it could even yield further improvements to other over-fitted BabyLMs, generating further performance improvements.

\section{Implementing LTG-Baseline Pre-training}
\label{sec:hyper}
We must define the exact specifics of training our own LTG-Baseline model. This requires us to prepare the corpus, define a training script, and specify the exact hyperparameters. Following the specifications from the authors \citep{hu2024findings}, we use the predefined BabyLM corpus established in Section~\ref{sec:corpora-main}, and adopt the LTG architecture by \citet{samuel2023trained} as the starting point. However, an issue with further implementation is that the authors have not published the LTG-Baseline training script or its hyperparameters.

To resolve this, we reached out directly to the authors, who supplied us with all the necessary details. They used the standard MLM training script by \citet{wolf-etal-2020-transformers} with hyperparameters, which are provided in Table~\ref{tab:hyperparams}.
\begin{table}[h!]
  \centering
  \begin{tabular}{@{}l|l@{}}
    \toprule
    \textbf{Hyperparameter} & \textbf{Value} \\
    \midrule
    max\_seq\_length              & 128 \\
    per\_device\_train\_batch\_size & 128 \\
    num\_train\_epochs            & 20 \\
    learning\_rate                & 5e-4 \\
    adam\_beta1                   & 0.9 \\
    adam\_beta2                   & 0.999 \\
    adam\_epsilon                 & 1e-8 \\
    max\_grad\_norm               & 1 \\
    warmup\_steps                 & 0 \\
    \bottomrule
  \end{tabular}
  \caption{\label{tab:hyperparams}LTG-Baseline pre-training hyperparameters}
\end{table}

Furthermore, for purposes of model comparison, epochs cease to be a representative unit when we alter the corpus. To remedy this, we note that, in the standard training, 20 epochs translate to roughly 140,000 training steps. Thus, we compare all of our models on the interval from 0 to 140,000 steps, removing any unfair advantage stemming from an expanded corpus.

\section{LLM-in-the-loop Debiasing Implementation}
\label{sec:LLM}
The LLM-in-the-loop detoxification approach has garnered interest because it allows us to remove toxicity without removing information from a corpus \citep{yuan2025llm}. The method uses an LLM to rewrite any toxic sentence in a non-toxic way whilst preserving the sentence’s meaning. Thus, we are able to remove toxicity without disrupting coherence.

\begin{figure*}[!t]
    \centering
    \includegraphics[width=0.75\linewidth]{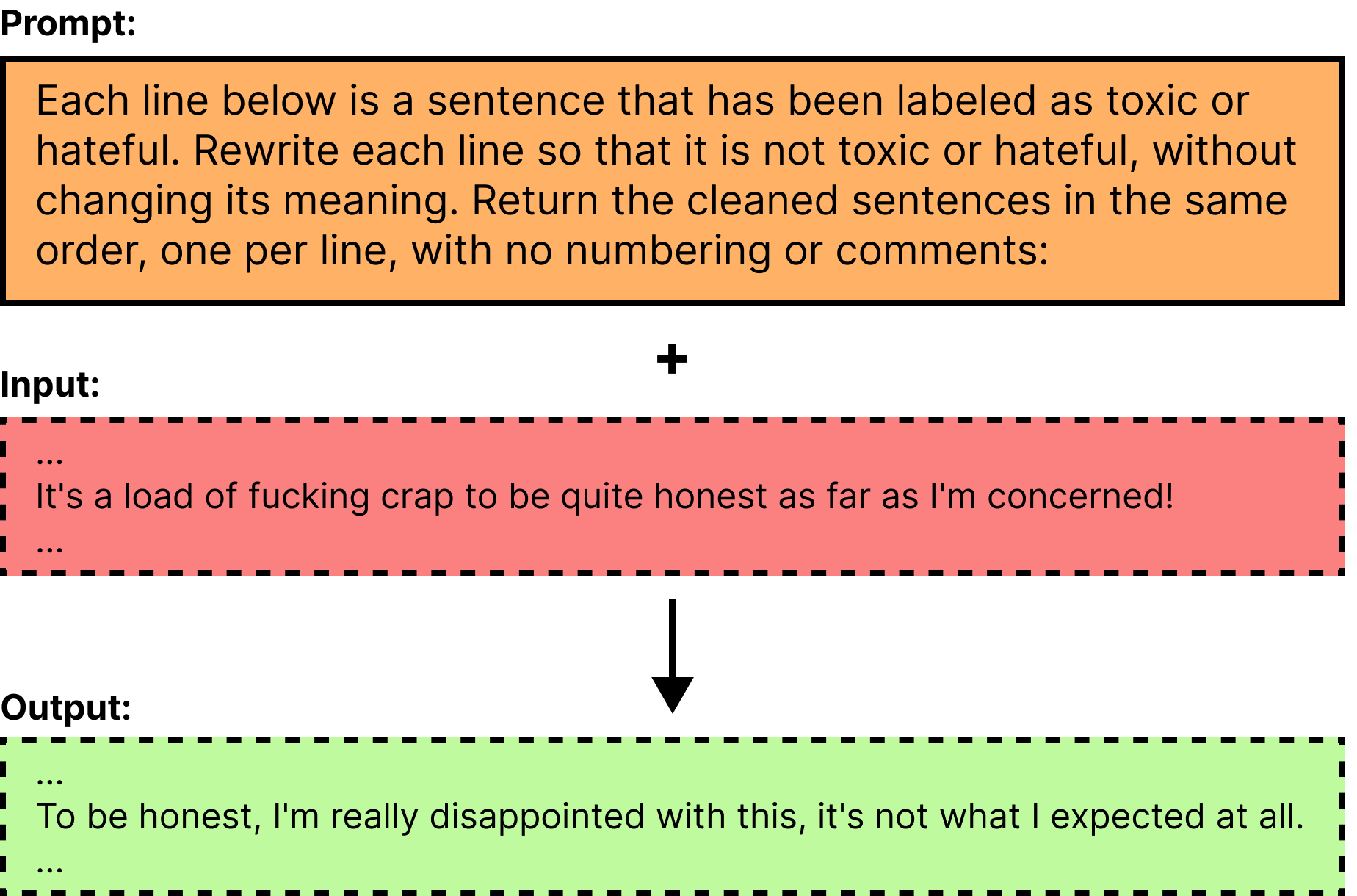}
    \caption{\label{fig:prompt}Detoxification prompt and its effect on an example}
\end{figure*}

For the purposes of implementation, we selected Llama-3.3-70B to detoxify the sentences \citep{grattafiori2024llama}. It was picked due to being a widely used and high-performing open-source model. In order to identify a well-performing detoxification prompt, we sampled 20 toxic sentences and tested multiple prompt variants, with the selected one, together with an example, shown in Figure~\ref{fig:prompt}.

\section{Perturbation Augmentation Implementation}
\label{sec:perturb-implement}
The Perturbation Augmentation debiasing technique by \citet{qian-etal-2022-perturbation} relies on the idea of using a pre-trained LLM (perturber) to rewrite the corpus, randomly swapping every demographic reference for a different one. 

\begin{figure*}[h!]
    \centering
    \includegraphics[width=0.75\linewidth]{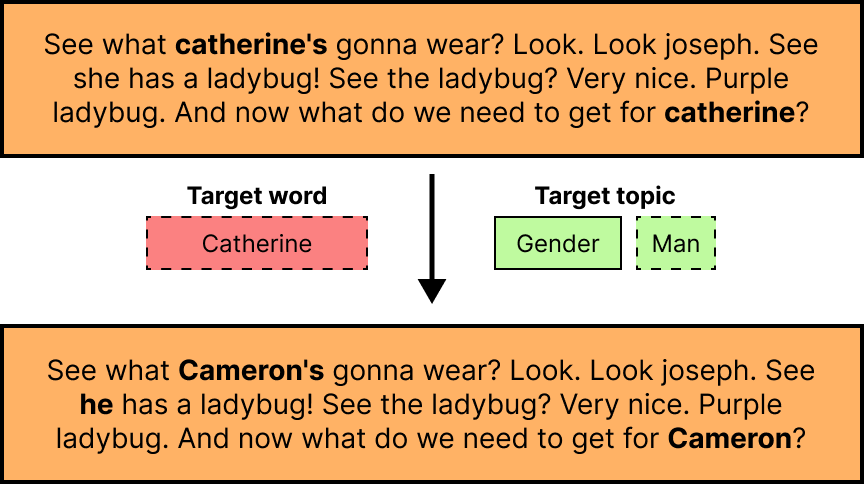}
    \caption{\label{fig:per-graph}Perturbation showcase}
\end{figure*}

In practice, this means that when supplied with a chunk of text, a target word, and a target topic, the perturber changes the gender, age, or race of the subject in the chunk to a new one randomly selected from the set list in Figure~\ref{tab:per-data}. If applied to the entire corpus, this ensures uniform distribution of topics. An example of this kind of perturbation is shown in Figure~\ref{fig:per-graph}.

To apply this perturbation to the BabyLM corpus, we use the perturber trained by the authors, ensuring the quality of the result. Unfortunately, we also need the target words, which the authors have not provided. Nevertheless, we resolved this by extracting a noisier list of target words from the perturber’s training data.

During the perturbation process, we split the corpus into 128-token chunks, extract all target words from the chunk, randomly select one of them together with a random sub-category, and perturb the chunk. We only focus on race and gender perturbations as age proved to generate more erroneous changes. If the chunk does not change, we repeat this process with another word until we change the chunk or exhaust the target words. Thus, the noisy ineffective words do not disrupt the experiment. With the entire corpus transformed, Table~\ref{tab:per-data} details the distribution of changes.

\begin{table}[htbp]
\centering
\footnotesize
\setlength{\tabcolsep}{3pt}
\begin{tabularx}{0.5\textwidth}{l|l|c}
\toprule
\textbf{Category} & \textbf{Subcategory} & \textbf{Perturbed Chunks} \\
\midrule
\multirow{3}{*}{Overall Corpus}
  & Any Change & 84.9\% \\
  & Gender     & 79.0\% \\
  & Race       & 5.9\%  \\
\cmidrule(lr){1-3}
\multirow{3}{*}{Gender Perturbation}
  & Non-binary & 27.9\% \\
  & Woman      & 26.8\% \\
  & Man        & 24.2\% \\
\cmidrule(lr){1-3}
\multirow{6}{*}{Race Perturbation}
  & Pacific-Islander & 1.1\% \\
  & Native-American  & 1.0\% \\
  & White            & 1.0\% \\
  & Asian            & 1.0\% \\
  & Black            & 1.0\% \\
  & Hispanic         & 1.0\% \\
\bottomrule
\end{tabularx}
\caption{\label{tab:per-data}Change distribution in the perturbed corpus}
\end{table}
\section{Generative AI}
An AI assistant was used to help with grammar and language editing.
\section{Computational Resources}
For all debiasing experiments, covering both fine-tuning and pre-training, we utilised a server with four A100 GPUs. Most evaluation tasks were conducted on a single NVIDIA T4 GPU. In total, including test runs, the experiments consumed approximately 700 GPU-hours. No hyper-parameter search experiments were conducted for any of the experiments.

\section{Code Repository}
The code is available in a dedicated GitHub repository.\footnote{\url{https://github.com/trhlikfilip/bias-dynamics-sandbox}}

\end{document}